\documentclass{article}
\usepackage{arxiv}
\usepackage[utf8]{inputenc} 
\usepackage[T1]{fontenc}    
\usepackage{hyperref}       
\usepackage{url}            
\usepackage{booktabs}       
\usepackage{amsfonts}       
\usepackage{nicefrac}       
\usepackage{microtype}      
\usepackage{lipsum}		
\usepackage{graphicx}
\usepackage{natbib}
\usepackage{doi}

\usepackage{amsmath,amssymb,amsfonts}
\usepackage{mathrsfs}
\usepackage{calrsfs}

\usepackage{amsthm}

\newtheorem*{remark}{Remark}

\newtheorem*{assumption*}{\assumptionnumber}
\providecommand{\assumptionnumber}{}
\makeatletter
\newenvironment{assumption}[2]
 {%
  \renewcommand{\assumptionnumber}{Assumption #1}%
  \begin{assumption*}%
  \protected@edef\@currentlabel{#1}%
 }
 {%
  \end{assumption*}
 }
\makeatother

\usepackage{algorithmic}
\usepackage{textcomp}

\title{Design, Modelling and Control of an Amphibious Quad-Rotor for Pipeline Inspection}

\author{ 
	\href{https://orcid.org/0000-0003-2701-9257}{\includegraphics[scale=0.06]{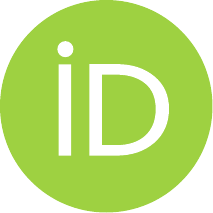}\hspace{1mm}Petar Durdevic} \\
	Department of Energy Technology\\
	Aalborg University\\
	Esbjerg, Denmark 6700 \\
	\texttt{pdl@energy.aau.dk} \\
	\And
	\href{https://orcid.org/0000-0002-2934-3803}{\includegraphics[scale=0.06]{orcid.pdf}\hspace{1mm}Shaobao Li} \\
	Department of Energy Technology\\
	Aalborg University\\
	Esbjerg, Denmark 6700 \\
	\texttt{shl@et.aau.dk} \\
	\And
	\href{https://orcid.org/0000-0002-1297-3702}{\includegraphics[scale=0.06]{orcid.pdf}\hspace{1mm}Daniel O.~Arroyo} \\
	Department of Energy Technology\\
	Aalborg University\\
	Esbjerg, Denmark 6700 \\
	\texttt{doa@energy.aau.dk} \\
}

\renewcommand{\shorttitle}{\textit{arXiv} Template}

\hypersetup{
pdftitle={A template for the arxiv style},
pdfsubject={q-bio.NC, q-bio.QM},
pdfauthor={David S.~Hippocampus, Elias D.~Striatum},
pdfkeywords={First keyword, Second keyword, More},
}

\begin{document}
\maketitle

\begin{abstract}
	Regular inspections are crucial to maintaining waste-water pipelines in good condition. The challenge is that inside a pipeline the space is narrow and may have a complex structure. The conventional methods that use pipe robots with heavy cables are expensive, time-consuming, and difficult to operate.  
In this work, we develop an amphibious system that combines  a quad-copter with a surface vehicle, creating a hybrid unmanned aerial floating vehicle (HUAFV). 
Nonlinear dynamics of the HUAFV are modeled based on the dynamic models of both operating modes. The model is validated through experiments and simulations. 
A PI controller designed and tuned on the developed model is implemented onto a prototype platform. Our experiments demonstrate the effectiveness of the new HUAFV's modeling and design.
\end{abstract}


\section{Introduction}

Denmark has around 80,400 km of sewerage pipes connecting 6$\%$ of its land area \cite{MSTKloak,VanditalDanva2019} and renovating this system was estimated to cost (150M USD) \cite{VanditalDanva2019} in 2018. Moreover, poorly maintained pipes can lead to high groundwater infiltration, which is a significant problem in Denmark. An official report estimates the infiltration to be anywhere between 150-400$\%$ of the sewage, which is around 150-200M m$^3$ of infiltrated water. This number is expected to rise due to climate change \cite{VanditalDanva2019}. The consequence is a high hydraulic load on the wastewater treatment plants, leading to reduced treatment efficiency. 

%

Sewer pipelines require periodic inspections and early fault detection to prevent groundwater infiltration. 
In the last decades, an extensive variety of robotic systems have been developed to perform inspections due to their increasingly lower cost and flexibility. 
In general, robotic systems for in-pipe inspections can be classified as pig robot type, wheeled, caterpillar, wall-press, walking, inchworm, or screw type robots \cite{Roh2007}.
Among all robotic solutions, cable-tethered or remotely operated vehicles (ROV) wheeled robots are most commonly used, \cite{nassiraei2007concept,ahrary2007study}.
Wheeled robotic systems have limitations, in spite of their simplicity. 
For instance, they can operate only in dry pipes and may not be able to crawl up in pipeline structures with high inclinations, or in ill-constrained pipelines that are broken.
One of the problems that ROVs encounter in the pipelines of city sewage systems is sand sediments that tend to accumulate over time making ROV's movement difficult. 
Due to this, sewage inspections in certain municipalities in Denmark are performed by closing off the pipeline under consideration. Additionally, the sand and debris must be washed out with water and the water has to be drained. This operation requires abundant amounts of water, time, and energy.
To address the limitations that crawling and submersible robots have, Vertical Takeoff and Landing (VTOL) and Unmanned Air Vehicles (UAV) capable of flying inside the pipelines have emerged. These commercial semi-autonomous drones such as Elios 2 have been used to inspect London city sewage \cite{london} and Barcelona \cite{FLYABILITY_2018}.
However, flying inside a pipeline poses great challenges due to its internally constrained space. But since the drone needs to keep hovering over the water's surface, the flight time is reduced.

Some commercial designs such as the Splash-drone \cite{SplashDrone,lloyd2017evaluation}, are capable of flying and landing in water. The Aqua-copter, the QuadH20 and the Mariner Quad-copter are drones that are designed to fall into the water, \cite{quadh2o,mariner_drone}. 
%
The aforementioned systems, however, are designed for manual piloting, and thus not suitable for flying inside pipelines where wireless signals could be blocked.

Other autonomous solutions such \cite{aquadrone} have been designed to monitor waterways, ports, and sea.
%
%
This paper presents the design and modeling of a fully autonomous HUAFV, which we call the \textit{\textbf{Quad-float}}. This platform is efficient for sailing in the pipelines, flying to access difficult areas, or being deployed down a man-hole when required.

%
 %

To our knowledge, no research work has addressed the problem of designing HUAFVs capable of navigating inside confined areas, such as pipelines, nor on the control and navigation of HUAFVs on water surfaces. 

A survey of early hybrid systems is presented in \cite{yang2015}. Other research articles have discussed other more recent types of hybrid systems called: Hybrid Unmanned Aerial Underwater Vehicles (HUAUV), Hybrid Aerial Underwater Vehicles (HAUV), and unmanned aerial underwater vehicles (UAUV) as was reported in  \cite{drews2014hybrid,neto2015attitude,da2018comparative,maia2017design,lu2019adaptive,mercado2019aerial,zha2019towards,ma2018research,ZubiMR18,Ma2018}.

The Loon Copter described in \cite{ZubiMR18} is an autonomous quadcopter with active buoyancy control that is capable of performing aerial, water-surface, and subaquatic diving. A closed-loop control system is used to perform aerial and water-surface missions and an open-loop control system is used for diving.
A similar Hybrid Unmanned Aerial Underwater Vehicle is presented in \cite{Ma2018} where a dynamic model of the transition from air to underwater media process and its control system is developed and tested by simulation.   

In \cite{lu2019adaptive} a hybrid aerial underwater vehicle (HAUV) is considered, which uses four rotors both for underwater and air navigation. 
The model is extended with time-varying added mass and damping, additionally, it is assumed that the restoring torques are zero, i.e. $\tau_R = 0$, since the center of buoyancy and gravity coincide.
Similar ideas have been explored in the field of HUAUV, \cite{drews2014hybrid,neto2015attitude,da2018comparative,maia2017design,mercado2019aerial,zha2019towards}, where the restoring torques are not considered, and only the restoring forces due to buoyancy are considered. The reason for this is that these works focus on submersible vehicles.
In our work, the restoring torques are crucial due to surface-dwelling, as the movement in $x$ and $y$ is generated by changing the pitch and roll angle respectively, and thus the restoring torque dynamic is crucial. The restoring torques are modeled following the meta-critic restoring forces \cite{fossen1994guidance}.


The Quad-float presented in this paper is a novel amphibious drone aimed at navigating autonomously inside pipelines, capable of flying and navigating by floating on the water's surface. However, this work addresses only the modeling issues for the flotation paradigm.
A nonlinear dynamic model of the platform was developed based on quad-copter rigid body kinematics, which includes the drag forces and torques and the restoring forces and torques associated with floating vehicles. A novelty of this Quad-float is the addition of the effect of restoring torques in its model.
A PID controller was designed to control the position of the vehicle while it is floating.
A prototype of the Quad-float was built and used for model validation and control implementation, using inbuilt LiDAR measurements for vehicle localization. 
%
%
\section{The Quad-float Concept}
\begin{figure}[h]
	\centering		\includegraphics[width=0.90\columnwidth]{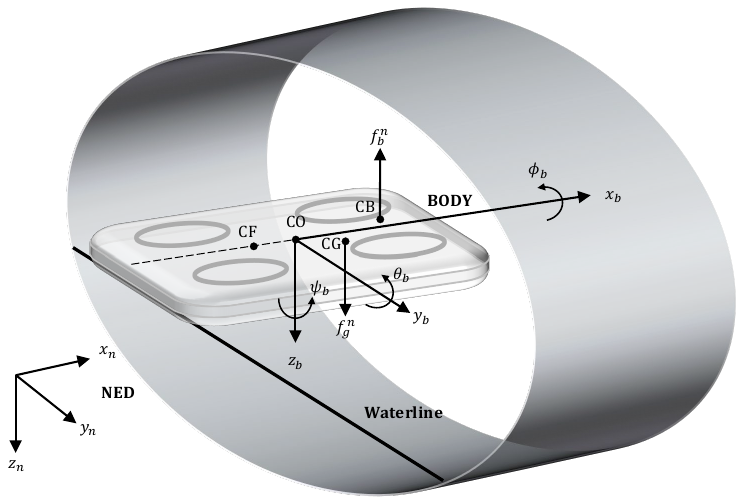}
	\caption{Sketch of the Quad-float in a pipeline environment floating in a layer of water, with the relevant kinematic and kinetic parameters and forces.}
	\label{fig:Capture}
\end{figure}
The Quad-float is an amphibious vehicle, that combines a quad-copter and a flotation device. During the flight, the Quad-float has the same six degrees of freedom (DOFs) as the quad-copter.
%
When in water, it moves like a boat, which is normally only modeled with 3 DOFs \cite{fantoni1999stabilization}. However, in the relatively small Quad-float, the rotation in roll and pitch cannot be neglected due to the nonlinear couplings between the four rotors. Therefore, a 6-DOF model will also be used to describe the motion in water. 

In this paper, an important addition to the standard quad-copter model is the inclusion of buoyancy and thus meta-critic stability and hydrodynamic forces.  
During the flight, a quad-copter movement is affected by aerodynamic forces, torques, and the gravitational force $f_g$. However, the aerodynamic forces can be neglected due to the relatively low velocity during flight.

\begin{figure}[h]
	\centering
		\includegraphics[width=0.5\columnwidth]{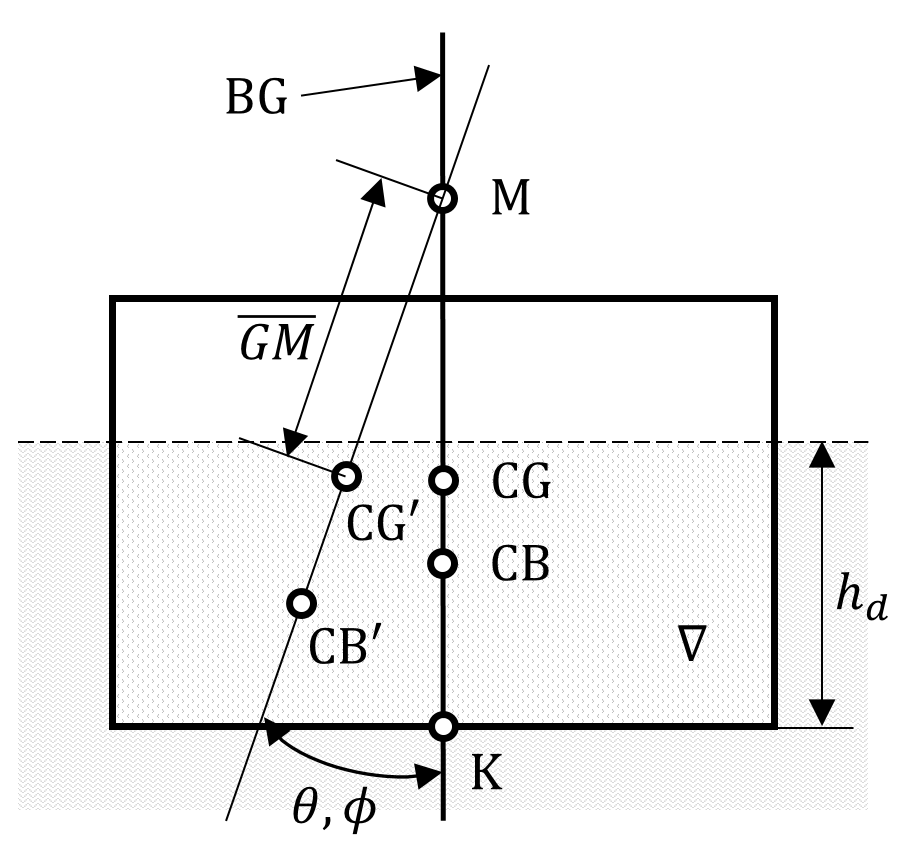}
	\caption{Meta-critic Stability}
	\label{fig:Metacritic_stability}
\end{figure}

\begin{remark}
A crucial design feature of the Quad-float is that the center of buoyancy (\textbf{CB}), must be placed below the center of gravity (\textbf{CG}) to achieve meta-critic stability, as shown in figure \ref{fig:Metacritic_stability}. 
\end{remark}

It is expected that when the Quad-float lands in the water, the hydrodynamic and static forces and torques will be dominant, similarly to those affecting a flotation vehicle.
Figure \ref{fig:Capture} illustrates the Quad-float when floating on water inside a pipeline.
%
\section{Designed System}
The proposed amphibious platform consists of three major parts: a frame attached to a pontoon and the electronics equipment housed in two waterproof compartments. 
%
%
\subsection{Frame}
The Quad-float follows the \textbf{X} type design \cite{zhang2014survey}. This setup was chosen as it gives us the largest actuation force for relatively low individual motor rotation velocity. 
%
%
\subsection{Pontoon}
The pontoon is made from a 50 mm Styrofoam sheet, which has been cut into the shape of the quad-rotor's frame.
%
\subsection{Control and Propulsion System}
The Quad-float is equipped with four Emax RS2205-2600KV brushless DC motors (BLDC), controlled by an Airbot typhoon 32 v2 $4 \times 35$ A electronic speed controller (ESC).
%
A 10 DOFs inertial measurement unit (IMU) is used (Bosch BNO055).
An ST VL53L0X LiDAR is used for relative $x$ and $y$ distance measurements
An NXP MK66FX1M0VMD18 Kinetis K66: 180 MHz Cortex-M4F micro-controller unit (MCU) is used as the processing unit
%
%
%
\section{Dynamic Model of the Quad-Float}
%

Several general assumptions are made to simplify the modeling problem. 
\begin{assumption}{1}
-The vehicle is symmetrical with a symmetrical mass distribution.
\end{assumption}
\begin{assumption}{2}
-The motor dynamics are fast in comparison to system dynamics and can be ignored.
\end{assumption}
\begin{assumption}{3}
-As the vehicle travels at a relatively low speed in the water, the added mass can be neglected.
\end{assumption}
\begin{assumption}{4}
-The water surface is considered to be flat and without waves.
\end{assumption}
\begin{remark}
Assumptions 1-3 are valid in general. Regarding Assumption 4, experiments were done in a flat water surface such that the wave disturbance induced by the motion of the Quad-float was relatively small and thus can be neglected. This simplification was made to reduce the complexity of the model. It is worth mentioning that in a real environment wave disturbances may affect the Quad-float due to water flow. To address this problem a disturbance estimator should be designed for disturbance rejection. This work will be done in a future paper.
\end{remark}

\subsection{Rigid Body Kinematics}
A quad-copter can be described as a rigid body, with 6 DOFs.
The generalized coordinates of a quad-copter are $\zeta=[x,y,z,\phi,\theta,\psi]^T \in \mathbb{R}^6$ defined in the inertial frame, where $p = [x,y,z]^T$ denotes the position referred to surge, sway, and heave, respectively and $\eta = [\phi,\theta,\psi]^T$ denotes the Euler angles referred to roll, pitch, and yaw, respectively.
The kinetic and potential energies, $(K,P)$, represented in the generalized coordinates can be computed through the Lagrangian:
\begin{equation}
	L(\zeta,\dot{\zeta}) = K-P	= K_{trans} + K_{rot} - P,
\label{eq:eq:euler_lagrange_1}
\end{equation}
where the potential energy is defined as $P=- m g_z p_z + G \zeta$, where $m$ is the mass of the vehicle, $g_z$ is the gravitational acceleration and $G \zeta$ are the restoring forces.
The translational kinetic energy is defined as $K_{trans} = \frac{1}{2}m\dot{p}^T\dot{p} $, and the rotational kinetic energy is defined as $K_{rot} = \frac{1}{2}\omega_b^T I \omega_b$, in which $\omega_b$ is the angular velocity resolved in the body fixed frame $(*_b)$, and $I$ is the inertia tensor
%
which due to the symmetry of the vehicle can be a diagonal matrix defined as:
\begin{equation}
I = diag[I_{xx},I_{yy},I_{zz}].
\label{eq:inertia_symetric_1}
\end{equation}
The angular velocity $\omega_b$ is related to the generalized coordinates angular velocity vector $\eta$ through a kinematic relationship:
\begin{equation}
	\dot{\eta} = W^{-1}(\eta) \omega_b
\label{eq:generalized_relationship_1}
\end{equation}
where $W$ is the Euler angle transformation matrix from the body frame to the inertial frame, refer to equation \ref{eq:transformation_matrix_euler_1}, using the $[z,y,x]$ convention \cite{fossen1994guidance,castillo2004stabilization}:
\begin{equation}
	W= \begin{bmatrix}
	1 & 0 &-s_{\theta}  \\
	0 & c_{\phi} &   c_{\theta} s_{\phi}\\
	0 & -s_{\phi} & c_{\theta} c_{\phi}
	\end{bmatrix},
\label{eq:transformation_matrix_euler_1}
\end{equation}
%
where $c_{*}=cos(*)$ and $s_{*}=sin(*)$.
The inertia tensor can then be represented in the generalized coordinates, i.e. $\eta$, as:
\begin{equation}
	 \mathbb{J} =  \mathbb{J} (\eta) = W^T I W
\label{eq:generalized_inertia_1}
\end{equation}
leading to:
\begin{equation}
	K_{rot} = \frac{1}{2}\dot{\eta}^T \mathbb{J} \dot{\eta},
\label{eq:Kinetic_rotational_1}
\end{equation}
which gives the final Lagrangian
\begin{equation}
\begin{aligned}
	L =  K-P 
		=  \frac{1}{2}m\dot{p}^T\dot{p} + \frac{1}{2}\dot{\eta}^T \mathbb{J} \dot{\eta} - m g_z p_z + G\zeta
	\end{aligned}
\label{eq:euler_lagrange_2}
\end{equation}
\subsection{Euler-Lagrange}
After applying the Euler-Lagrange equation we get the full dynamics with external generalized forces:
\begin{equation}
	\frac{d}{dt} \frac{\partial L}{\partial \dot{\zeta}} -\frac{\partial L}{\partial \zeta} = \begin{bmatrix}
	F_p\\
	\tau
	\end{bmatrix}
\label{eq:Euler_lagrange_full_dynamics_1}
\end{equation}
where the translational force $F_p=FR\in R^3$ with $R$ being rotational matrix, defined by equation (\ref{eq:rotation_matrix}), \cite{choi2014nonlinear,goldstein2002classical}. Here $F=[0, 0, T_{tot}]$ is referred to as the translational force with $T_{tot}$ as the total thrust generated by the four propellers, and $\tau = [\tau_{\phi},\tau_{\theta},\tau_{\psi}]$ the generalized moments for the attitude control. 
F and $\tau$ will be discussed in section \ref{sec:motor_external_forces}.
%
%
\begin{equation}
R = \begin{bmatrix} 
  c_{\psi}  c_{\theta} &  c_{\psi}  s_{\phi}  s_{\theta}- c_{\phi}  s_{\psi} &  s_{\phi}  s_{\psi}+ c_{\phi}  c_{\psi}  s_{\theta}\\
 c_{\theta}  s_{\psi} &  c_{\phi}  c_{\psi}+ s_{\phi}  s_{\psi}  s_{\theta} &  c_{\phi}  s_{\psi}  s_{\theta}- c_{\psi}  s_{\phi}\\ 
- s_{\theta} &  c_{\theta}  s_{\phi} &  c_{\phi}  c_{\theta}
\end{bmatrix}
\label{eq:rotation_matrix}
\end{equation}
As we have no cross-terms in the kinetic energy combining $\dot{P}$ and $\dot{\eta}$ from equation (\ref{eq:euler_lagrange_2}), the Euler-Lagrange equations can be divided into translational and rotational dynamics, \cite{garcia2006modelling}.
\begin{equation}
m\ddot{P} + mge^3 = F_p, \hspace{.5cm} \mathbb{J} \ddot{\eta} +  \dot{\mathbb{J}} \dot{\eta} - \frac{1}{2} \frac{\partial}{\partial \eta} (\dot{\eta}^T \mathbb{J} ) = \tau.
\label{eq:euler_lagrange_3}
\end{equation}
%
%
We can represent the Coriolis/Centripetal vector, $C(\eta, \dot{\eta})$, as is equation (\ref{eq:Coriolis_vector_1}), which includes the gyroscopic and the centrifugal terms \cite{choi2014nonlinear,castillo2005stabilization,garcia2006modelling,raffo2010integral}.\\
\begin{equation}
	C(\eta, \dot{\eta})  = \dot{\mathbb{J}} \dot{\eta} - \frac{1}{2} \frac{\partial}{\partial \eta} (\dot{\eta}^T \mathbb{J}).
\label{eq:Coriolis_vector_1}
\end{equation}
Thus, we can simplify the kinetics term from  (\ref{eq:euler_lagrange_3}) to:
\begin{equation}
	\mathbb{J} \ddot{\eta} + 	C(\eta, \dot{\eta})  = \tau.
\label{eq:simplify_kinetics_1}
\end{equation}
\subsection{Hydrodynamic Forces and Moments - Damping}

Damping is a non-conservative force and is added as an external force.
For a surface vehicle, damping is induced by a multitude of external effects, and the total hydrodynamic damping can be formulated as follows \cite{fossen2011handbook}:
\begin{equation}
	D(\dot{p}) \triangleq D_P(\dot{p}) + D_S(\dot{p}) + D_W(\dot{p}) + D_M(\dot{p})
\label{eq:damping_1}
\end{equation}
where $D_P(\dot{p})$, $D_S(\dot{p})$, $D_W(\dot{p})$ and $D_M(\dot{p})$ are the radiation-induced potential damping, linear and quadratic skin friction, wave drift damping, and vortex shedding damping, respectively \cite{fossen1994guidance}.
The hydrodynamic damping $D(\dot{p})$ affects the motion of the vehicle in a highly nonlinear and coupled fashion. Practically, it is not trivial to determine the higher-order terms and the off-diagonal terms in $D(\dot{p})$ \cite{fossen1994guidance}.
We, therefore, choose to simplify the damping $D(\dot{p})$ by assuming that it is non-coupled and thus we can simplify $D(\dot{p})$ to:
\begin{equation}
D=diag[D_X,D_Y,D_Z,D_K,D_M,D_N]
\label{eq:daming_final}
\end{equation}
where $[D_X,D_Y,D_Z,D_K,D_M,D_N]$ are the damping forces in the 6 DOFs.
Identification of the drag coefficients is, again, not trivial, and is in this work done by a trial and error method based on experience with the experimental setup, alternatively, the parameters can be identified from data.
$D$ can be represented as the translational and rotational parts as follows:
\begin{equation}
\begin{aligned}
D_p&=diag[D_X,D_Y,D_Z], \\
D_{\eta}&=diag[D_K,D_M,D_N].
\end{aligned}
\label{eq:daming_final_eta}
\end{equation}
\subsection{Hydrodynamic Forces and Moments - Restoring Forces and Moments}
Quad-float is a surface vehicle and will thus be affected by the same restoring forces as a ship in $[\phi, \theta, z]$. 
The vehicle is designed such that the weight and the buoyancy are in balance:
\begin{equation}
	mg=\rho g \nabla,
\label{eq:weight_buoyancy_balance_1}
\end{equation}
where $\nabla$ is the volume of the displaced fluid.
Due to restoring forces, the system will be open-loop stable in $[\phi, \theta, z]$, referred to as meta-critic stability.
The restoring forces are governed by the meta-centric height $M_i$ the center of buoyancy (CB) and the center of gravity (CG) \cite{fossen1994guidance,fossen2011handbook}.

The restoring forces in $z$ and the moments in $[\phi,\theta]$ can be written as follows \cite{fossen2011handbook}:
\begin{equation}
	\begin{aligned}
		Z_{restoring} & = -\rho g A_{wp} z, \\
		K_{restoring} & = -\rho g \nabla \overline{GM_T} sin \phi, \\
		M_{restoring} & = -\rho g \nabla \overline{GM_L} sin \theta. 
	\end{aligned}	
\label{eq:restoring_forces_1}
\end{equation}
where $A_{wp} = L  H$ is the water plane area, $\rho$ is the density of the displaced fluid, $g$ is gravity and $\overline{GM_T}$ and $\overline{GM_L}$ are the transverse and longitudinal meta-centric height (m), respectively (i.e. the distance between the meta-center $M_i$ and \textbf{CG}), refer to figure \ref{fig:Metacritic_stability}.
For surface vessels, we can use a linear approximation, following the assumptions in \cite{fossen1994guidance}:
\begin{itemize}
	\item $\phi$, $\theta$, $z$ $\approx 0$.
	\item $\int{}_0^z A_{wp} (z_\zeta) d z_\zeta \approx A_{wp} (0) z $.
	\item $sin(\theta) \approx \theta, cos(\theta) \approx 1$, $sin(\phi) \approx \phi, cos(\phi) \approx 1$. 
\end{itemize} 
Then we have $g(\zeta)\approx G\zeta$ and thus $g(\zeta)$ becomes:
%
%
%
\begin{equation}
	G=diag[0,0,\rho g A_{wp}(0),\rho g \nabla \overline{GM_T},\rho g \nabla \overline{GM_L},0],
\label{eq:restoring_final}
\end{equation}
where $G$ can be represented as the translational and rotational parts as follows:
\begin{equation}
\begin{aligned}
G_p&=diag[0,0,\rho g A_{wp}(0)],\\
G_{\eta}&=diag[\rho g \nabla \overline{GM_T},\rho g \nabla \overline{GM_L}, 0 ].
\end{aligned}
\label{eq:restoring_final_eta}
\end{equation}
\subsection{Vectorial Representation of the Dynamics}
The model in terms of translation and rotation forces can be structured as follows:
\begin{equation}
\begin{aligned}
m \ddot{p}&= F_p - m g_z  - D_p  \dot{p} - G_p p,\\
\mathbb{J} \ddot{\eta}&= \tau - C(\eta,\dot{\eta}) \dot{\eta} - D_{\eta}  \dot{\eta} - G_{\eta}  \eta.
\end{aligned}
\label{eq:Final_model}
\end{equation}

The addition of the restoring forces and moments $(G_p p,G_{\eta} \eta)$ and the damping forces and moments $(D_p \dot{p},D_{\eta} \dot{\eta})$ distinguish this model from the standard quad-rotor model. From a general control perspective, the model has been expanded to include the un-damped natural frequency $\omega_{n}$ and the damping ratio. In essence, this enables the presented system to be stable, unlike the standard quad-rotor definition, changing the control objective whilst in water.

\subsection{Motor External Forces} \label{sec:motor_external_forces}
The propulsion on a quad-rotor consists of four rotor-blades, with a constant pitch, that is attached to a Brush-Less Direct Current (BLDC) motor's axis.
Thus by alternating the angular speed of the motor $\omega_{i, i = 1,\cdots,4}$, the individual motor's thrust $T_{i, i = 1,\cdots,4}$ and torque $\tau_{i, i = 1,\cdots,4}$ are alternated.


This leads to the following generalized forces and moments with respect to $z_b$ and $\eta$.
\begin{equation}
	\begin{bmatrix}
		T_{tot}\\
		\tau
	\end{bmatrix} = \begin{bmatrix}
		T_{tot}\\
		\tau_{\phi}\\
		\tau_{\theta}\\
		\tau_{\psi}
	\end{bmatrix} = \begin{bmatrix}
	     \sum{}_{i=1}^{4} T_i \\
		 (-T_1-T_3+T_2+T_4) \cdot l_x/2 \\
		 (T_1+T_2-T_3-T_4) \cdot l_y/2 \\
		 (\tau_{m_1}-\tau_{m_2}+\tau_{m_3}-\tau_{m_4}) 
	\end{bmatrix}
	\label{eq:motor_total_1}
\end{equation}

\section{Experimental Model Validation}


To validate the model, experiments were performed in a pool with a diameter of 1.5 $m$, a wall height of 0.3 $m$, and a water depth of 0.08 $m$. 




Three 1.5 second impulse response experiments were performed for roll, pitch, and yaw with the following amplitudes: 0.1 $N/m$, 0.1 $N/m$, 0.005 $N/m$ respectively. 
These values were heuristically determined by performing some experiments, as they had a significant effect on the system. 
\begin{figure}[h]
	\centering
		\includegraphics[width=.90\columnwidth]{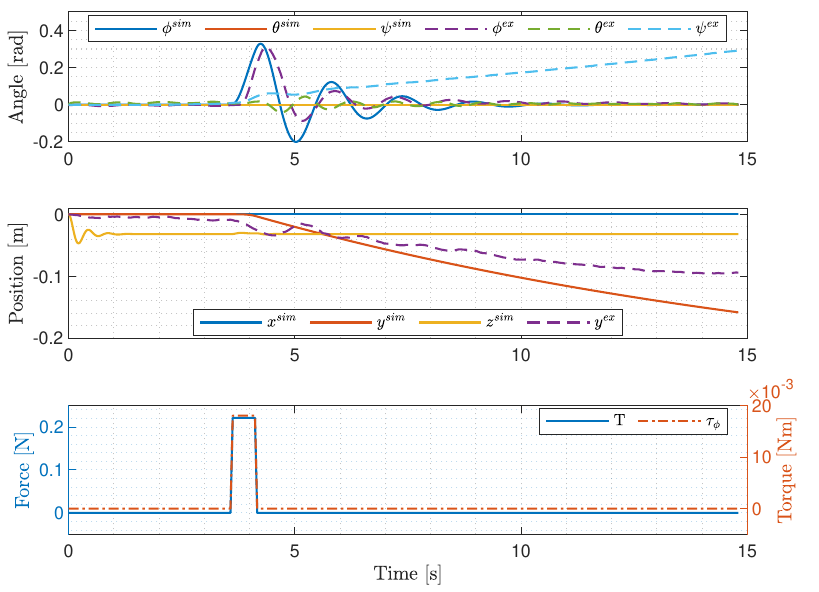}
	\caption{Model validation regarding rolling, where a step input on $\tau_{\phi}$ was applied.}
	\label{fig:Roll_val_1}
\end{figure}
\subsection{Validation Result Discussion}
\begin{figure}[h]
	\centering
		\includegraphics[width=.9\columnwidth]{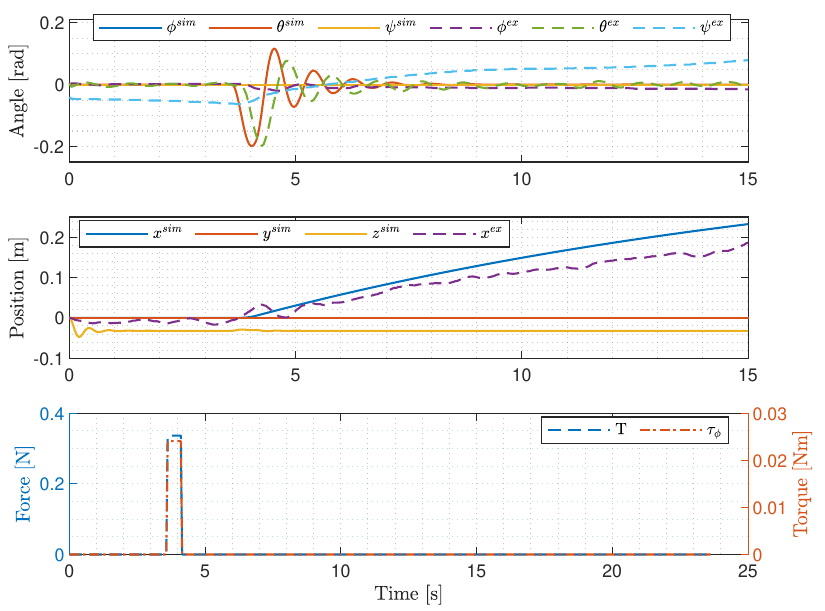}
	\caption{Model validation regarding pitching, where a step input on $\tau_{\theta}$ was applied}
	\label{fig:Pitch_val_1}
\end{figure}

The model performed well regarding the roll and pitch angles, as seen in figures \ref{fig:Roll_val_1}, \ref{fig:Pitch_val_1}, with a relatively good fit to the experimental data obtained. 
An input in roll torque influences the pitch and yaw, and vice-versa. This can be attributed to multiple causes: i) the complex nonlinear coupling in the system which is not represented by the model, ii) due to the motor's velocity and iii) the unsymmetrical nature of the physical structure of the platform. 
The translational performance of the model was evaluated using Lidar measurements, and as can be seen in the results, a steady state is not reached due to the size of the pool used in the experiments, which has only 25 cm of travel distance available.
Because the measurements are performed with a LiDAR, the distance is slightly offset due to the rotation in $\psi$. 
The model's fit to the experimental data regarding an input in $\tau_{\psi}$, is relatively good, with a small offset where the experimental data is changing faster than the simulation result. 
\begin{figure}[h]
	\centering
		\includegraphics[width=.9\columnwidth]{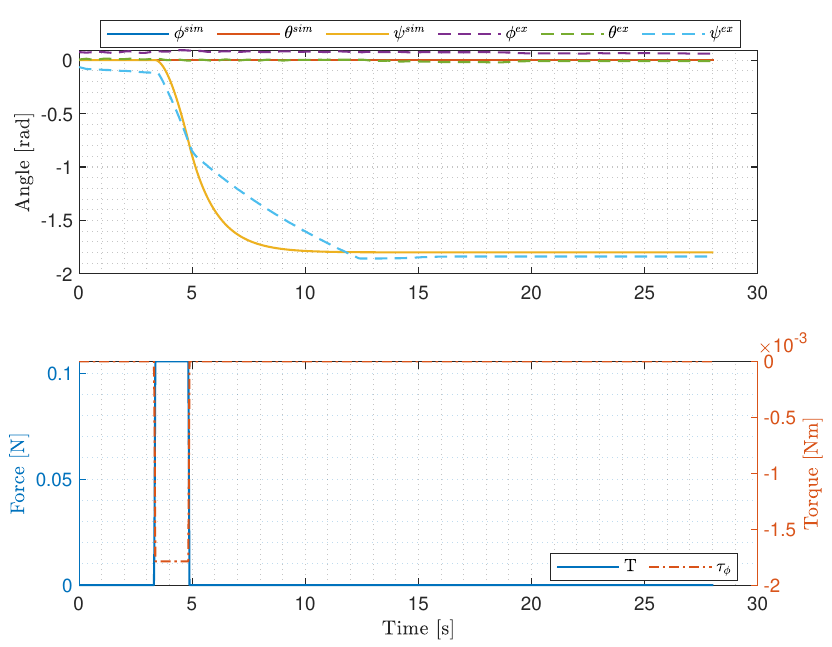}
	\caption{Model validation regarding yawing, where a step input on $\tau_{\psi}$ was applied}
	\label{fig:Yaw_val_1}
\end{figure}
\section{Control System Design}
Since the aerial flight control is no different from existing algorithms \cite{michael2010grasp}, in this work we focus only on the sailing control, where the quad-float moves like a boat.
When in water the Quad-float is open-loop stable in $[z,\phi,\theta]$, thus our objective is to control $\bar{p}=[x_d,y_d]$ and $\bar{\eta}=[\psi_d]$ and track the references $\bar{p}_d=[x_d,y_d]$ and $\bar{\eta}_d=[\psi_d]$. 
The Quad-float is under-actuated wrt. $[\bar{p}]$, in the same manner as the quad-copter and to achieve the translational movement in $\bar{p}$ the force $F_x$ and $F_y$ is required which is governed by non-linear coupling
\begin{equation}
\begin{aligned}
F_x&=T_{tot}\cdot(s(\phi)s(\psi) + c(\phi)c(\psi)s(\theta)),\\
F_y&=-T_{tot}\cdot(c(\psi)s(\phi) - c(\phi)s(\psi)s(\theta)).
\end{aligned}
\label{eq:nonlinforce}
\end{equation}
i.e. to obtain $F_x$ and $F_y$ requires the system to be actuated in $[\theta,\phi]$, which is achieved through an input in $[\tau_{\phi},\tau_{\theta}]$ following the system model, (\ref{eq:Final_model}).
In the system model from equation (\ref{eq:Final_model}) a rotational movement in $\psi$ is achieved through $[\tau_{\phi}]$.
The two feedback control loops are considered separately, i.e. the desired positions for $\bar{p}_d=[x_d,y_d]$ and $\bar{\eta}_d=[\psi_d]$. The control structure can be structured as follows:
\begin{equation}
\begin{aligned}
	\tau_{\bar{p}}&= [\tau_{\phi},\tau_{\theta}] = e_{\bar{p}} \cdot K_{{\bar{p}}},\\
	\tau_{\bar{\eta}} & = [\tau_{\psi}] = e_{\bar{\eta}} \cdot K_{{\bar{\eta}}}.
	\end{aligned}
\label{eq:control_eta}
\end{equation}
where $e_{\bar{p}} = \bar{p}_d - \bar{p}$ and $e_{\bar{\eta}} = \bar{\eta}_d - \bar{\eta}$ are the tracking errors regarding $\bar{p}$ and $\bar{\eta}$ respectively, and $K_*$ is the feedback control parameter.
Here we refer to $K_p$ as the translational controller i.e. with respect to $x$ and $y$, i.e. $K_x$ and $K_y$, and to $K_{\eta}$ as the rotational controller with respect to $\psi$, i.e. $K_{\psi}$. 

%
In the current work, the feedback control parameters $K_*$ are substituted by a proportional-integral controller.
\subsection{Controller Tuning}

During the early experiments with the platform, we experienced that the system would become unstable if it reached angles above 0.3 rad due to the physical position of the \textbf{CG} relative to the \textbf{CB} and the design of the Pontoon.
In order to keep the system stable the translational controller was tuned, such that the $\phi < 0.1 rad$ and $\theta < 0.1 rad$ for steps of 0.1 $m$.
Initially, we chose a rise-time of 2.5 $s$, but it was reduced to 5 $s$ due to issues with the physical setup. 
Although the $\psi$ actuation is decoupled from the $\phi$ and the $\theta$ actuation, the physical construction of the pontoons does not allow for aggressive $\psi$ actuation. 
Some experiential tests were performed to test the platform, where steps of 0.2 rad with a rise-time of 1 $s$ were found to be safe and used as a controller requirement.
The controller parameters were designed using Matlab's PID tuning toolbox.
%
%
\subsection{Control Simulation and Experimental Results}
In the current work, we consider two operating conditions: i) navigate within the pipeline in the translational plane $[x,y]$ to \textit{`locate the damage'}, ii) adjust the $\psi$ angle $[\psi]$ to \textit{`inspect the damage'}.
Two scenarios were chosen to evaluate the tracking performance of the controller regarding the position $y$ and angle $\psi$. 
Additionally, the same reference signals are used both in the simulations and in the experiments and are shown in equations (\ref{eq:y_step}) and (\ref{eq:psi_step}). 
\begin{equation}
y_d =
  \begin{cases}
    0 m    & \quad \text{if } t<t_{step}\\
    0.1 m  & \quad \text{if } t>t_{step}
  \end{cases}
\label{eq:y_step}
\end{equation}
\begin{equation}
\psi_d =
  \begin{cases}
    0 rad   & \quad \text{if } 0s<t<5s\\
    0.1745  rad & \quad \text{if } 5s<t<10s\\
		0.1745 \cdot 2 rad & \quad \text{if } 10s<t<15s\\
		\vdots
  \end{cases}
\label{eq:psi_step}
\end{equation}
\subsubsection{Simulation Results}
\begin{figure}[h]
	\centering
		\includegraphics[width=.9\columnwidth]{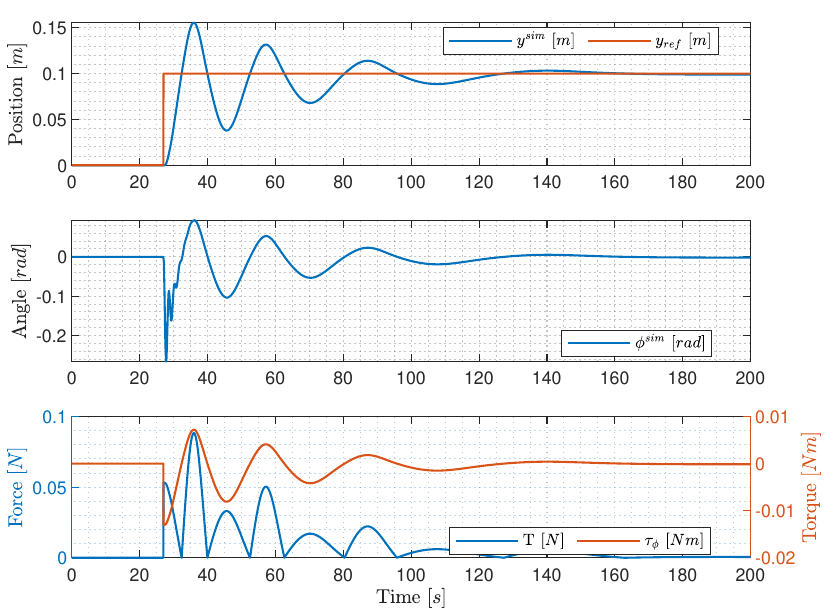}
	\caption{Simulation Results: Reference tracking performance of the controller regarding $y_d$.}
	\label{fig:Sim_xstep1}
\end{figure}

\begin{figure}[h]
	\centering
		\includegraphics[width=.9\columnwidth]{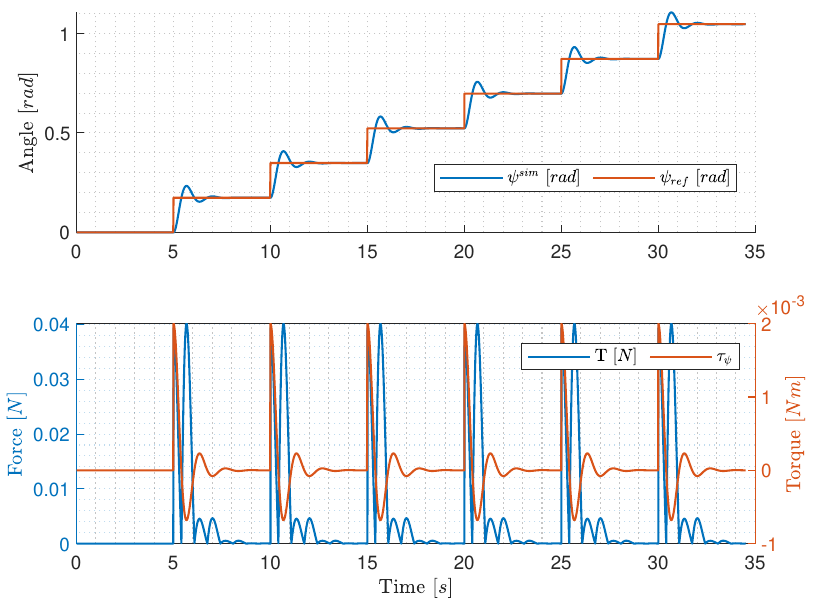}
	\caption{Simulation Results: Reference tracking performance of the controller regarding $\psi_d$.}
	\label{fig:Sim_psistep1}
\end{figure}

The simulation results are shown in figures \ref{fig:Sim_xstep1} and \ref{fig:Sim_psistep1}, regarding the step response in $y$ and angle $\psi$, respectively. 
The controller performs according to the design specifications in both cases.
\subsubsection{Experimental Results}
\begin{figure}[h]
	\centering
		\includegraphics[width=1.00\columnwidth]{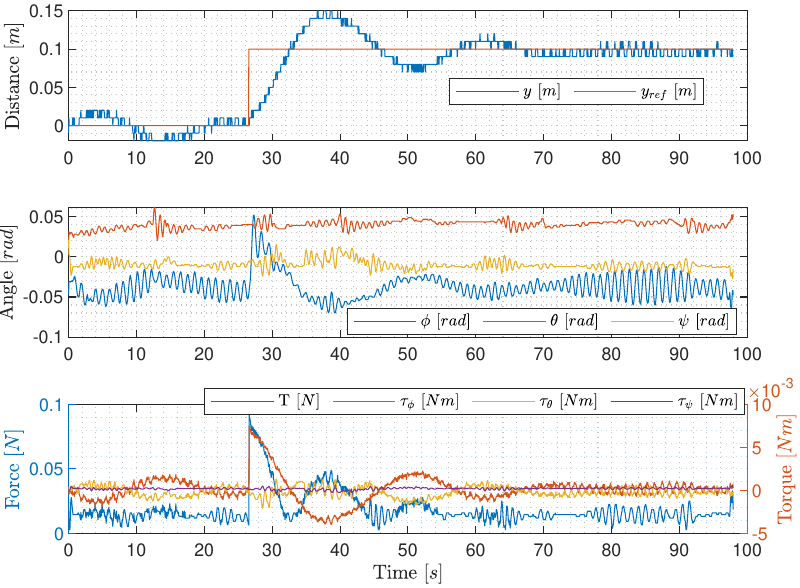}
	\caption{Experimental result: step test in $y$ direction, (the angle offset is not adjusted for tilt)}
	\label{fig:xstep1}
\end{figure}

\begin{figure}[h]
	\centering
		\includegraphics[width=1.00\columnwidth]{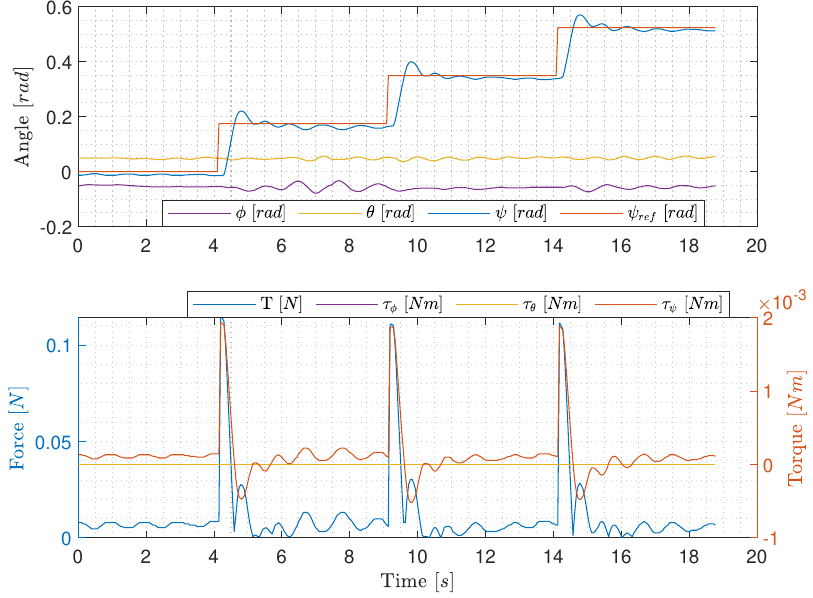}
	\caption{Experimental result: step test in $\psi$ direction}
	\label{fig:psistep1}
\end{figure}

The same controller parameters are implemented onto the hardware platform, and the controller was evaluated following the same reference signals, specified in equations (\ref{eq:y_step}) and (\ref{eq:psi_step}). The experiments were performed in the same pool as was used for the model validations, the experimental results are respectively displayed in figures \ref{fig:xstep1} and \ref{fig:psistep1}.
\begin{table}[h]
\caption{Control Comparison: simulation and experimental results}
\label{tab:controller_performance}
\begin{tabular}{l|l|l|l}
$y$              & $y_{sim}$ & $y_{exp}$ & $\%$ deviation \\ \hline
Rise-time   [0-100 $\%$]  &5.38 $s$& 5.1313 $s$ &  4.62 $\%$ \\
Peak-time     &8.9 $s$& 10.1373 $s$ &   13.9 $\%$\\
Peak    			&0.16 $m$& 0.15 $m$& 6.25 $\%$\\
Settling-time [2$\%$] &148 $s$	 & \textit{(71.35)} $s$& \textit{(51.79)} $\%$ \\
\midrule \hspace{.1cm}
$\psi$              & $\psi_{sim}$ & $\psi_{exp}$ & $\%$ deviation \\ \hline
Rise-time  [0-100 $\%$]   & 0.38 $s$& 0.4452 $s$ &  18.42 $\%$ \\
Peak-time     & 0.68 $s$& 0.6988 $s$ & 2.94 $\%$ \\
Peak    			& 0.26 $rad$& 0.2204 $rad$ &  15.38 $\%$ \\
Settling-time [2$\%$] & 3.75 $s$& \textit{(5.06)} $s$ &  \textit{(34.93)} $\%$ 
\end{tabular}
\end{table}
Controller performance measures are shown in table \ref{tab:controller_performance}, showing the rise-time, peak-time, peak value, and settling time for the simulation and experimental results. 
It is clear that the transient performance of experimental results is similar to the simulated results, indicating that the model is a good representation of the true system. 
In the simulations, the translational controller tracks the reference, and the $\phi$ angle is kept below 0.3 $rad$. In the experiment, the $\phi$ angle never reaches more than 0.1 $rad$, which is good for system stability and shows that the system has the potential to be more aggressive. 

In the case of the rotational controller, the simulations show a rise time of 0.38 $s=t_{100}$ and a settling time of 3.75 $s$, which are good performance measures considering that the system actuates 0.1745 $rad$ at each step. 
Similar behavior is seen in the experimental data only with a slightly slower response. During the experiments, it was observed that the actuation system was very powerful and could perform much more than is seen here.  

Note that the settling time of the translational controller was never reached due to the accuracy of the measurement, which is 0.01m.
Therefore the settling time, shown in the parenthesis, is the final value of $y$ in the experiment.
In addition, in the experimental results the rotational controller is affected by oscillatory behavior which results in it not settling at a steady state value, and thus the settling time, in parenthesis, is the final value of $\psi$ in the step, i.e. after 5 $s$. 
The rise-time, peak-time, and peak values regarding the rotational and the translational controller are similar in the experimental and simulation results, none deviating more than 20$\%$.

\section{Discussion}
\subsection{Model Discussion}
The comparison between the simulated model response to the experimental data collected from the designed platform, seen in figures \ref{fig:Roll_val_1},\ref{fig:Pitch_val_1} and \ref{fig:Yaw_val_1}, indicates that the model fits well with the experimental results. 
A few exceptions are listed below, together with potential causes.
\subsubsection{$\phi, \theta$}
The experimental data has an $\approx 0.18$ $s$ and $\approx 0.19$ $s$ delay for the two angles respectively, this is caused by the initial build-up of thrust from the propeller, this dynamic was not included in the model. The second observation is the oscillations in $\phi$ $\&$ $\theta$, this is caused by the waves in the pool and increases when the vehicle moves and creates additional waves in the pool which bounce off the pool wall.
The reason is that as the system is under-actuated, a movement in the translational directions $x$ and $y$ is created by altering the $\theta$ and $\phi$ angles respectively. This motion displaces the fluid below the pontoons and thus creates a trough after which follows a crest and the frequency of the platform's angular rotation around $x$ and $y$ will be translated into the wave frequency.
%
\subsubsection{$\psi$} The experimental data and the simulation have the same response at $\tau_{\psi} \neq 0$, but at $\tau_{\psi} = 0$ after 5 $s$ the experimental data is damped faster, both reach approximately the same steady-state value. This could be caused by coupling in the damping. 
\subsubsection{$x,y$} The simulated translational position follows the experimental data well, deviating as the distance grows. \footnote{The distance measurement is affected by the measurement method; where the distance to the wall is measured with the LiDAR, and as the platform changes its $\psi$ position the distance changes.}
\subsection{Controller Discussion}
The rotational controller's performance demonstrates the capability of the Quad-float for fast reference tracking. 
This is desirable for pipeline inspections as the orientation of the Quad-float for inspections is crucial. 
The experimental analysis of the controllers indicates that the disturbance has an impact on the controller's reference tracking as both the rotational and the translational controller are affected by oscillations.

The oscillations produced are body-induced waves, i.e. waves created by the movement of the body in the water pool and the reflections of the waves in the pool's wall. The oscillations can for example be seen in figure \ref{fig:xstep1} from 80 $s$ to 90 $s$ in the $\phi$ angle.
As a consequence, the system does not reach the 1 $\%$ steady state value during the experiments but instead oscillates with $\approx$ $10\%$ from the steady state. 
%
\section{Conclusion}
This work presents the design and testing of a hybrid unmanned aerial floating vehicle (HUAFV), named the Quad-float. The purpose of this vehicle is to provide a platform that can navigate pipelines flooded with water by floating on the water's surface
%
A dynamic model of the Quad-float has been developed and experimentally validated, on the constructed platform. 
The model was used as a simulation platform for testing and tuning a PI control structure, which was later implemented on the platform. 
Experiments were performed to test the controller's translational and rotational tracking performance.
The translational tracking performance was achieved according to design specifications and is limited by the hardware design. Improvements in the hardware design, to achieve a larger pitch and roll angle, will allow for a larger translational force and thus a faster system response.
As the concept presented in the paper is new, there are several challenges that should still be solved and thus there is potential for improvement. 
In the current work only the control system for moving on the water surface was developed, future work will explore designing a controller for taking off and landing on a moving surface within a confined environment. 






\bibliographystyle{unsrtnat}
\bibliography{bib.bib}  






\end{document}